\documentclass{article}

\usepackage{PRIMEarxiv}

\usepackage[utf8]{inputenc} 
\usepackage[T1]{fontenc}    
\usepackage{hyperref}       
\usepackage{url}            
\usepackage{booktabs}       
\usepackage{amsfonts}       
\usepackage{nicefrac}       
\usepackage{microtype}      
\usepackage{lipsum}
\usepackage{fancyhdr}       
\usepackage{graphicx}       
\graphicspath{{media/}}     

\usepackage{times}
\usepackage{epsfig}
\usepackage{graphicx}
\usepackage{booktabs}
\usepackage{cite}
\usepackage{amsmath,amssymb,amsfonts}
\usepackage{algorithmic}
\usepackage{graphicx}
\usepackage{textcomp}
\usepackage{xcolor}
\usepackage{soul}
\usepackage{etoolbox}
\usepackage{enumitem}
\usepackage{wrapfig}
\usepackage{pifont}

\usepackage[toc,page]{appendix}
\usepackage{caption}
\usepackage[caption=false]{subfig}

\title{Data-Side Efficiencies for Lightweight Convolutional Neural Networks}

\author{Bryan Bo Cao$^{\dagger\ddagger}$\thanks{Work done during an internship at Nokia Bell Labs.}, \hspace{6pt} Lawrence O’Gorman$^{\ddagger}$, \hspace{6pt} Michael Coss$^{\ddagger}$, \hspace{6pt} Shubham Jain$^{\dagger}$\\
$^{\dagger}$ Stony Brook University, Stony Brook, NY, USA\\
$^{\ddagger}$ Nokia Bell Labs, Murray Hill, NJ, USA \\
{\tt\small $^{\dagger}$\{boccao,jain\}@cs.stonybrook.edu,\hspace{6pt}$^{\ddagger}$\{bryan.cao\}@nokia.com,
}\\
{\tt\small $^{\ddagger}$\{larry.o\_gorman,mike.coss\}@nokia-bell-labs.com 
}
}

\begin{document}
\maketitle

\begin{abstract}
\label{sec:abstract}
We examine how the choice of data-side attributes for two important visual tasks of image classification and object detection can aid in the choice or design of lightweight convolutional neural networks. We show by experimentation how four data attributes -- number of classes, object color, image resolution, and object scale affect neural network model size and efficiency. Intra- and inter-class similarity metrics, based on metric learning, are defined to guide the evaluation of these attributes toward achieving lightweight models. Evaluations made using these metrics are shown to require $30\times$ less computation than running full inference tests. We provide, as an example, applying the metrics and methods to choose a lightweight model for a robot path planning application and achieve computation reduction of $66\%$ and accuracy gain of $3.5\%$ over the pre-method model.
\end{abstract}



\vspace{-10pt}
\section{Introduction}
\label{sec:introduction}


Traditionally for computer vision applications, an algorithm designer with domain expertise would begin by identifying handcrafted features to help recognize objects of interest. More recently, end-to-end learning (E2E) has supplanted that expert by training a deep neural network to learn important features on its own. Besides the little forethought required about data features, there is usually only basic preprocessing done on the input data; an image is often downsampled and converted to a vector, and an audio signal is often transformed to a spectogram. In this paper, we use the term ``data-side'' to include operations that are performed on the data before input to the neural network. Our proposal is that a one-time analysis of data-side attributes can aid the design of more efficient convolutional neural networks (CNNs) for the many-times that they are used to perform inferences.


On the data side of the neural network, we examine four independent image attributes and two dependent attributes, the latter which we use as metrics. The independent attributes are \textbf{number of classes, \textbf{object color}, image resolution} and \textbf{object scale}. The metrics are \textbf{intra-} and \textbf{inter-class similarity}. Our goal is to optimize the metrics by choice of the independent variables -- specifically to maximize intra-class similarity and minimize inter-class similarity -- to obtain the most computationally efficient model.

Unlike benchmark competitions such as ImageNet \cite{ImageNet2015}, practical applications involve a design stage that can include adjustment of input specifications. In Section \ref{sec:relatedwork}, we tabulate a selection of applications. The ``wildlife'' application reduced the number of animal and bird classes from 18 to 6 in the Wildlife Spotter dataset~\cite{applWildlifeAustralia2017}. In the ``driving'' application \cite{applDriving2021}, the 10 classes of the BDD dataset \cite{applBDDdataset2018} were reduced to 7 by eliminating the ``train'' class due to few labeled instances
and combining the similar classes of rider, motor, and bike into rider.



The main contributions of this paper are:
\begin{enumerate}
\vspace{-5pt}
\item Four data-side attributes are identified, and experiments are run to show their effects on the computational efficiency of lightweight CNNs.\vspace{-6pt}
\item Intra- and inter-class similarity metrics are defined to aid evaluation of the four independent attribute values. Use of these metrics is shown to be about $30\times$ faster than evaluation by full inference testing.\vspace{-6pt}
\item Procedures are described using the similarity metrics to evaluate how changing attribute values can reduce model computation while maintaining accuracy.\vspace{-6pt}
\item Starting with the EfficientNet-B0 model, we show how our methods can guide the application designer to smaller ``sub-EfficientNets'' with greater efficiency and similar or higher accuracy.
\end{enumerate}

We describe related work in Section \ref{sec:relatedwork}. Each of the attributes is defined in Section \ref{sec:metrics}, procedures are described to apply these toward more efficient models in Section \ref{sec:Method}, and experimental evidence is shown in Section \ref{sec:experiments}. We conclude in Section \ref{sec:conclusion}.


\section{Related Work}
\label{sec:relatedwork}

\label{SectionBG}

The post-AlexNet \cite{alexNet2012} era (2012-) of convolutional neural networks brought larger and larger networks with the understanding that a larger model yielded higher accuracy (e.g., VGGNet-16 \cite{vggNet2014} in 2014 with 144M parameters). But the need for more efficiency, especially for embedded systems \cite{autonomousCarsMIT2023}, encouraged the design of lightweight neural networks \cite{sigProcSurvey}, such as SqueezeNet \cite{iandola2016squeezenet} in 2017 with 1.2M parameters. Models were reduced in size by such architectures as depthwise separable convolution filters ~\cite{MobileNet}. More efficient handling of data was incorporated by using quantization ~\cite{gysel2018ristretto, han2015deep, leng2018extremely}, pruning ~\cite{cheng2017survey, blalock2020state, StructPrune_li2016, shen2022structural}, and data saliency \cite{yeung2016end}. This model-side efficiency quest is a common research trend where new models are evaluated for general purpose classification and object detection on public benchmarks such as ImageNet ~\cite{ImageNet2015}, COCO ~\cite{lin2014microsoft}, and VOC ~\cite{everingham2009pascal}. Orthogonal and complementary to these model-side efficiencies ~\cite{redmon2016you, glenn_jocher_2021_5563715, wang2022yolov7}, we examine efficiencies that can be gained before the model by understanding and adjusting the data attributes within the confines of the application specifications. Early work ~\cite{kang2017noscope} optimizes models specialized to the target video only for binary-classification. Our work extends to multi-class classification and object detection. 

In Table \ref{tab:applications}, we list a selection of 9 applications whose data attributes are far less complex than common benchmarks. For these applications, class number is often just 2. The largest number of classes, 7, is for the ``driving'' \cite{applBDDdataset2018} application. Compare these with 80 classes for the COCO dataset and 1000 for ImageNet. For 2 applications, color is not used. For the ``crowd'' application, it is not deemed useful and for the ``ship, SAR'' application, the input data is inherently not color. The resolution range is not broad in this sampling, likely due to matching image size to model input width. Many papers did not describe the scale range; for these, we approximated from the given information or images in the paper. The broadest scale range (as a fraction of image size) is the ``driving'' application ($1/32$ to $1/2$), and the narrowest is for the ``mammals'' application, using aerial image capture, with scale from $1/30$ to $1/20$.

\begin{table}[h]
  \centering
  \begin{tabular}{ccccc}
   \toprule
    Application & $N_{Cl}$ & $N_{Co}$ & $R_{E}$ & $S_{C}$ \\
    \midrule   
    crowd \cite{appl_crowd} & 2 & no & $320 \times 240$ & $1/8$, $1/2$\\
    ship, SAR \cite{appl_shipSAR} & 2 & no & $416 \times 416$ & $1/20$, $1/6$\\
    
    cattle \cite{appl_cattle}& 2 & yes & $224 \times 224$ & $1/16$, $1/2$\\
    hardhat \cite{appl_hardhat} & 2 & yes & $300 \times 300$ & $1/25$, $1/2$\\
    
    wildlife \cite{applWildlifeAustralia2017} & 6 & yes & $224 \times 224$ & $1/6$, $1/2$\\
    PCB defect \cite{appl_PCB}& 6 & yes & $640 \times 640$ & $1/30$, $1/15$\\
    ship \cite{appl_shipOptical} & 6 & yes & $416 \times 416$ & $1/20$, $1/2$\\
    mammals \cite{applAfricanMammals2022} & 6 & yes & 2k $\times$ 2k & $1/30$, $1/20$\\
    
    driving \cite{applDriving2021} & 7 & yes & 608 $\times$ 608 & $1/32$, $1/2$\\
    \bottomrule
  \end{tabular}
  \caption{Examples of data attributes for object detection applications with attributes: number of classes ($N_{Cl}$), color ($N_{Co}$), input resolution ($R_{E}$), and scale range ($S_{C}$) as a fraction of image size.}
  \label{tab:applications}
\end{table}

We use a measure of class similarity to efficiently examine data attributes, based on neural network metric learning.
This term describes the use of learned, low-dimensional representations of discrete variables (images in our case). The distance between two instances in the latent space can be measured by L1 ~\cite{koch2015siamese}, L2, or cosine similarity ~\cite{metriclearning}. Previous studies ~\cite{veit2017conditional, vasileva2018learning, tan2019learning} focus on learning a single similarity latent space.
Differences between classification ~\cite{zhai2018classification} and ranking based losses ~\cite{hadsell2006dimensionality} have been studied in ~\cite{kobs2021different}. PAN incorporates attributes to visual similarity learning ~\cite{mishra2021effectively}. In Sections \ref{sec:Metrics_intrasim} and \ref{sec:Metrics_intersim} we extend this line of research by adapting the metric from ~\cite{metriclearning} to measure intra- and inter-class similarity to serve efficiency purposes.


\emph{In contrast to research to improve model performance on public benchmarks, our goal is to develop an empirical understanding of the effects of these attributes on common CNNs, and from this to provide practical guidelines to obtain lightweight CNNs in real-world scenarios.} Our use of intra- and inter-class similarity metrics enables an efficient methodology toward this goal. Practical, low-complexity applications as in Table \ref{tab:applications} can benefit from our investigation and method.




\section{Data Attributes and Metrics}
\label{sec:metrics}

This work is centered on the hypothesis  that, the easier the input data is to classify, the more computationally efficient the model can be at a fixed or chosen lower accuracy threshold. In this section, we describe each of the data attributes and their relationships to the hypothesis. We define metrics to obtain the dependent variables. And we describe procedures to adjust the independent variables to obtain metrics that guide the design of an efficient model.

We first introduce the term, Ease of Classification $EoC$ and hypothesize the following relationship exists with the data-side attributes,
\begin{equation}
\footnotesize
    \mathrm{EoC} \leftarrow ({S_1}, \frac{1}{S_2}) \leftarrow ( \frac{1}{N_{Cl}}, \frac{1}{N_{Co}}, \frac{1}{R_E}, \frac{1}{S_C}).
     \label{eqn:EoC}
\end{equation}




\noindent The symbol ($\leftarrow$) is used to describe the direct relationships in which the left expressions are related to the right. $EoC$ increases with intra-class similarity $S_1$ and decreases with inter-class similarity $S_2$. The dependent variable $S_1$ is related to the reciprocals of the independent variables, number of classes $N_{Cl}$, number of color channels $N_{Co}$, image resolution $R_E$, and object scale $S_C$. The dependent variable $S_2$ is directly related to these independent variables. The model designer follows an approach to adjust these independent variables to obtain similarity measurements that achieve high $EoC$. Note that we will sometimes simplify $N_{Cl}$ to $CL$ for readability in figures.

In Section \ref{sec:experiments} we perform experiments on a range of values for each attribute to understand how they affect model size and accuracy. However, these experiments cannot be done independently for each attribute because some have dependencies on others. We call these \textit{interdependencies} because they are 2-way. We discuss interdependencies below for two groups, \{$S_C$, $R_E$\} and \{$S_1$, $S_2$, $N_{Cl}$\}.

\subsection{Number of Classes, $N_{Cl}$}
\label{sec:Metrics_nClasses}
The $N_{Cl}$ attribute is the number of classes being classified in the dataset. Experimental results with different numbers of classes are shown in Section \ref{sec:Exp_numClasses}. In Section \ref{sec:applRobot} we present results of changing the number of classes for a robot path planning application.

\subsection{Object Colors, $N_{Co}$}
\label{sec:Metrics_colors}

The $N_{Co}$ attribute is the number of color axes, either 1 for grayscale or 3 for tristimulus color models such as RGB (red, green, blue). When the data has a color format, the first layer of the neural model has 3 channels. For grayscale data input, the model has 1 input channel. In Section \ref{sec:Exp_color}, we show the efficiency gain for 1 versus 3 input channels.

\subsection{Image Resolution, $R_E$}
\label{sec:Metrics_resolution}
\begin{figure}[t]
\centering
    \includegraphics[width=0.6\linewidth]{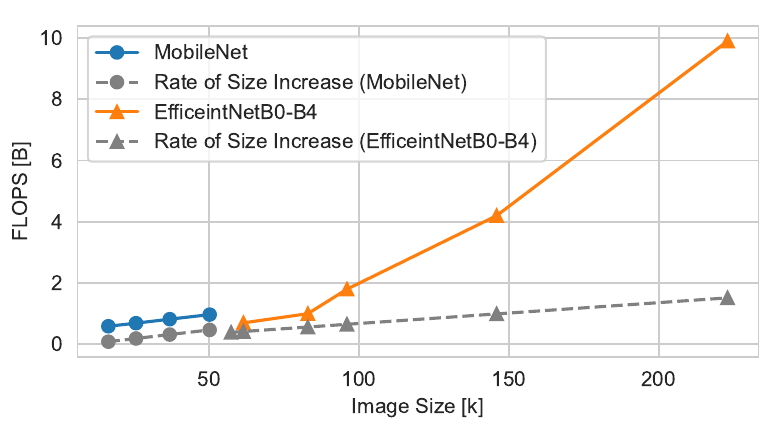}
   \caption{As the input image size is increased, the computation for the smaller MobileNet models increase at a similar rate. For the larger EfficientNet models (only up to EfficientNet-B4 are shown for clarity of display), the computation increases at a higher rate.}
   \label{fig:efficientNet_computation}
\end{figure}

Image resolution, measured in pixels, has the most direct relationship to model size and computation. Increasing the image size by a multiple $m$ (rows $I_r$ to $m \times I_r/2$ and columns $I_c$ to $m \times I_c$) increases the computation by at least one-to-one. In Figure \ref{fig:efficientNet_computation}, MobileNet computation increases proportionally to image size. The lower sized EfficientNet-B0 and -B1 models also increase proportionally, but rise faster with larger models B2, B3, and B4.

It is not the case that higher resolution always yields better accuracy. It usually plateaus, and that plateau is different for different classes involved. The objective is to choose the minimum image resolution before which accuracy drops unacceptably. Note that resolution and scale are dependent attributes, and neither can be adjusted without considering the effect on the other. Experimental results with different resolutions are shown in Section \ref{sec:Exp_res}.

\subsection{Object Scale, $S_C$}
\label{sec:Metrics_scale}
For a CNN, the image area, or receptive field, of a $3 \times 3$ filter kernel expands as the following sequence for layers $L=1, 2, 3, ..., L$,
\begin{equation}
\footnotesize
    3 \times 3, 7 \times 7, 15 \times 15, \ldots, (2^{L+1}-1) \times (2^{L+1}-1).
     \label{eqn:scaleFltrs}
\end{equation}
For object detection, this means that, if the maximum object size in an image is, for example, $15 \times 15$, the network needs at least 3 layers to yield features that convolve with (or overlap) the full size of these objects. In practice, from the sequence in equation \ref{eqn:scaleFltrs} the minimum number of layers can be found from the maximum sidelength $b_{max}$ of bounding boxes of objects over all image instances,
\begin{equation}
\footnotesize
    L \ge \lceil \mathrm{log}_2(b_{max} + 1) \rceil - 1.
     \label{eqn:scaleMinLayers}
\end{equation}
where $\lceil x \rceil$ is a ceiling operator, which rounds the $x$ value to the higher integer. For example, if $b_{max}$ is 250, the minimum number of layers needed is $ \lceil 7.966 \rceil - 1 = 7$.

In practice, the maximum object size is approximated as the longer sidelength of the bounding box of the object. In terms of model size and minimizing computation, by measuring the bounding box size of objects in a dataset, one can discover the maximum number of layers needed for full-object feature extraction. Furthermore, using filter kernels that are larger than the maximum object size increases extractions of inter-object features. When generalized to a sequence of rectangular filters, these tend toward Gaussian filters with increasing convolution layers, so the filter edge versus middle magnitude lowers as well.

\begin{figure}[t]
\centering
    \includegraphics[width=0.3\linewidth]{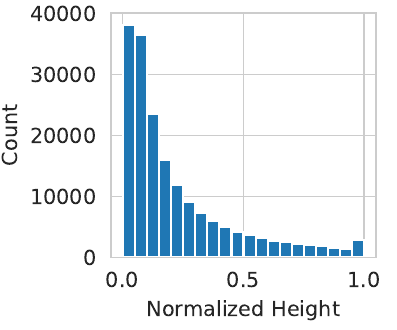}
    \hspace{40pt}
    \includegraphics[width=0.3\linewidth]{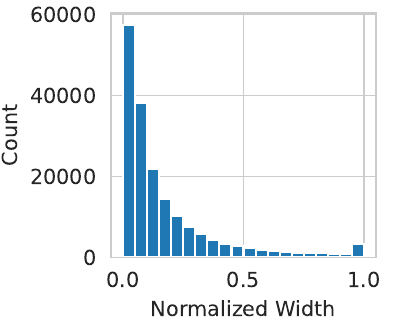}
    \caption{Histogram of bounding box sizes in COCO training dataset highly skewed at small sizes $< 0.2$.}
   \label{fig:scalePctOfImg}
\end{figure}

To decouple scale from resolution, we define the scale of an object as a fraction of the longer image dimension. With this definition, one can measure the scale of an object in an image without knowing the resolution. The \textit{size} of the object is measured in pixels; it is the product of scale times the longer dimension of the resolution. Figure \ref{fig:scalePctOfImg} shows the range of scales for objects in the COCO datase. Although 50\% of instances account for $\leq10\%$ of image size, still the sizes range from very small to full image size. Conversely, in many applications, scale range is known and much more contained as for some applications in Table \ref{tab:applications}. Experimental results with different scales are shown in Section \ref{sec:Exp_scale}.


\subsection{Resolution and Scale Interdependency}
\label{sec:scaleResolution}

The interdependence between the attributes \{$S_C$, $R_E$ \} is common to any image processing application, and can be described by two cases. For case 1, if the scale of objects within a class in an image is large enough to reduce the image size and still recognize the objects with high accuracy, then there is a resulting benefit of lower computation. However, reducing the image size also reduces resolution, and if there is another class that depends upon a high spatial frequency feature such as texture, then reducing the image would reduce overall accuracy. For case 2, if resolution of the highest frequency features of a class is more than adequate to support image reduction, then computational efficiency can be gained by reducing the image size. However, because that reduction also reduces scale, classes that are differentiated by scale will become more similar, and this situation may cause lower accuracy. We can write these relationships as follows,
\begin{equation}
\footnotesize
    (S_C \: \propto \: R_E ) \: \rightarrow \: \frac{1}{\mathrm{EoC}} ,
     \label{eqn:dependencies2}
\end{equation}
where the expression within parentheses is commutative (a change in $S_C$ will cause a proportional change in $R_E$, and vice versa), and these are inversely related to $EoC$.

\subsection{Intra-Class Similarity, $S_1$}
\label{sec:Metrics_intrasim}

Intra-class similarity is a measure of visual similarity between members of the same class as measured with vectors in the embedding space of a deep neural network. It is described by the average and variance of pairwise similarities of instances in the class,
\begin{equation}
\footnotesize
    S_1(C_1) = \frac{1}{N}\sum_{i,j \in C_1}\cos(\textbf{Z}_{i} \textbf{Z}_{j}), 
     \label{eqn:S1}
\end{equation}
\begin{equation}
\footnotesize
    \sigma_{S1}^2(C_1) = \frac{1}{N}\sum_{i,j \in C_1}(S_{ij} - S_1)^2, 
     \label{eqn:sigma_S1}
\end{equation}
\noindent where $C_1$ is a class containing a set of instances, $i$ and $j$ are indices in set $C_1$, $i \neq j$, and $N$ is the total number of similarity pairs $S_{ij}$ of two different instances in the class. $\textbf{Z}$ is the latent vector in the embedding space from a neural network trained by metric learning on instances of the class as well as other classes. (That is, it is the same model trained on the same instances as for inter-class similarity.) This metric is adapted from ~\cite{metriclearning}. We show the use of $S_1$ and $\sigma_{S1}^2$ in Section \ref{sec:SelectNCl}.

\subsection{Inter-Class Similarity, $S_2$}
\label{sec:Metrics_intersim}

Inter-class similarity is a measure of visual similarity between classes as determined by the closeness of instances of two classes in the embedding space of a deep neural network. For 2 classes, it is defined as the average of pairwise similarities of instances between classes,
\begin{equation}
\footnotesize
    S_2(C_1, C_2) = \frac{1}{N}\sum_{i \in C_1, j \in C_2}\cos(\textbf{Z}_{i} \textbf{Z}_{j}), 
     \label{eqn:S2}
\end{equation}
\noindent where $C_1$ and $C_2$ are instance sets of two different classes, $i$ and $j$ are indices in sets $C_1$ and $C_2$ respectively, $N$ is the total number of pairs of two instances in two different classes, and $\textbf{Z}$ is the latent vector in the embedding space from a neural network trained by metric learning on instances that include both classes as well as other classes if there are more than 2.


For an application involving more than two classes, we choose the inter-class similarity measure to be the maximum of inter-class similarity measures over all class pairs,
\begin{equation}
\footnotesize
    \hat{S_2}(\{ C_K \}) = \max \{ S_2 (C_m, C_n) \},
     \label{eqn:S2hat}  
\end{equation}
where $\{C_K \}$ is the set of all classes for $0 <= k < K$, and $\{ (C_m, C_n) \}$ is all class pairs for $0 <= m,n < K$, $m \neq n$. 
We choose the maximum of similarity results of class pairs because maximum similarity represents the worst case -- most difficult to distinguish -- between two classes. Alternatively, one could calculate the average of inter-class similarities for each pair, however the operation of averaging will hide the effect of one pair of very similar classes among other dissimilar classes, and this effect is worse for larger $N_{Cl}$. We show the use of $\hat{S_2}$ in Section \ref{sec:SelectNCo}.

We also use a measure that is the normalized difference between the maximum and the average,
\begin{equation}
\footnotesize
    \Delta{S_2}(\{ C_K \}) = \frac{\hat{S_2} - S_2}{S_2}.
     \label{eqn:S2delta}  
\end{equation}
A larger $\Delta{S_2}$ indicates a higher value of worst case $\hat{S_2}$, so we seek low $\Delta{S_2}$ in the methods described in Sections \ref{sec:SelectNCl} and \ref{sec:SelectRS}.

\subsection{Intra- and Inter-Class Interdependency}
\label{sec:interIntraClass}

There is a strong interdependence between $S_1$ and $S_2$ with a secondary dependence upon $N_{Cl}$, as we will describe.
With other factors fixed, smaller intra-class similarity increases inter-class similarity because more heterogeneous classes with wider attribute ranges are being compared, thus reducing $EoC$. As $N_{Cl}$ is increased, this effect is exacerbated because there is higher probability of low $S_1$ and high $S_2$ pairs, and (similarly) because we use worst-case maximum in equation \ref{eqn:S2hat}. We can write these dependent relationships as follows,
\begin{equation}
\footnotesize
    (S_1 \propto 1 / S_2 ) \: \rightarrow \mathrm{EoC} \: , \: \mathrm{for } N_{Cl} \geq 2,
     \label{eqn:dependencies1}
\end{equation}
where the expression within brackets is commutative (either $S_1$ or $S_2$ can cause inverse change in the other, and the relationship becomes stronger as $N_{Cl}$ increases.


\section{Method}
\label{sec:Method}

The general approach toward finding the most efficient model is to select ranges of the independent attribute values \{$N_{Cl}$, $N_{Co}$, $R_E$, $S_C$\}, calculate the similarity measures \{$S_1$, $S_2$\} for selected values in the range, and choose those independent attribute values that minimize $S_2$ and maximize $S_1$. Step-by-step procedures for selecting each of the independent attributes are described below. These procedures should be followed in the order of the sections below.

\subsection{Selection of $N_{Cl}$}
\label{sec:SelectNCl}
 
If the application permits adjustment of the number of classes, then the following procedure can help finding class groupings and an associated $N_{Cl}$ that supports a more efficient model. Initialize $\Delta{S_2}=\infty$, and follow these steps,
\begin{enumerate}[label=\roman*]
  \vspace{-6pt}
  \item Choose class groupings.
  \vspace{-6pt}
  \item Calculate $\Delta{S_2}$ from equation \ref{eqn:S2delta}. If this is less than the current smallest $\Delta{S_2}$ then this grouping is the most efficient so far.
  \vspace{-6pt}
  \item If $\Delta{S_2}$ is low, one can choose to exit, or continue.
  \vspace{-6pt}
  \item If one decides to continue, calculate $S_1$ from equation \ref{eqn:S1} and $\sigma_{S1}^2$ from equation \ref{eqn:sigma_S1} for each class. The value of $S_1$ is for guidance to help understand why the current $S_2$ is good or bad (low or high) to give guidance on choosing the next groupings. In general, class grouping with high $S_1$ and low $\sigma_{S1}^2$ will yield higher accuracy. Repeat these steps.
\end{enumerate}
We want to stop when inter-class similarity is low, indicated by a low value of $\Delta{S_2}$. However there is subjectivity in this step due to the manual choice of groupings and because different applications will have different levels of intra-class homogeneity and inter-class distinguishability.
In practice, the procedure is usually run for a few iterations to understand relative $\Delta{S_2}$ values for different groupings, and the lowest is chosen.

\subsection{Selection of $N_{Co}$}
\label{sec:SelectNCo}
The application can gain computational advantage if \textit{all} classes can be reduced to grayscale; if all classes cannot be reduced, then the model must handle color and there is no computational advantage. Following are the steps to choose grayscale or color,

\begin{enumerate}[label=\roman*]
  \item For all classes in color and grayscale, calculate $\hat{S_2}$.
  \vspace{-6pt}
  \item If $\hat{S_2}$ for grayscale is less than or equal to $\hat{S_2}$ for color, choose a grayscale model and grayscale for all instances of all classes.
\end{enumerate}

If the procedure above does not result in the choice of grayscale, there is a second option. This requires more testing and does not directly yield computation reduction, but may improve accuracy.
\begin{enumerate}[label=\roman*]
\item For each class, calculate $\hat{S_2}$ against every other class for these four combinations of the $(C_1, C_2)$ pairs: (grayscale, grayscale), (grayscale, color), (color, grayscale), and (color, color).
\vspace{-6pt}
\item For each class $C_1$ whose $\hat{S_2}$ for (grayscale, grayscale) and (grayscale, color) is smaller than for (color, grayscale) and (color, color), choose to use grayscale for the $C_1$ class.
\end{enumerate}

\subsection{Selection of $R_E$ and $S_C$}
\label{sec:SelectRS}

We first adjust the attribute $R_E$ to find the lower bound of resolution with respect to acceptable accuracy. Initialize $\Delta{S_2}=\infty$, and follow these steps,
\begin{enumerate}[label=\roman*]
 \item Calculate $\Delta{S_2}$ from equation \ref{eqn:S2delta}. If this is less than the current smallest $\Delta{S_2}$ then this resolution is the most efficient so far.
 \vspace{-6pt}
  \item If $\Delta{S_2}$ is low, one can choose to exit, or continue.
  \vspace{-6pt}
  \item Reduce the resolution by half and repeat these steps.
\end{enumerate}
In practice, a few iterations is run to see when $\Delta{S_2}$ rises, then choose the resolution where it is lowest.

After the resolution has been chosen, maximum object scale is multiplied by resolution to find the maximum object size in pixels. Equation \ref{eqn:scaleMinLayers} is used to find an estimate of the lower bound of number of layers needed in the model.


\section{Experiments}
\label{sec:experiments}

In this section, we perform experiments on the data attributes to show how their values relate to model efficiency and accuracy. 
Note that our level of experimentation is not even across all attributes. We believe that the experiments upon number of classes, intra- and inter-class similarities, and resolution are sufficient to support the Ease of Classification relationship of equation \ref{eqn:EoC}. For color, we give a quantitative comparison of color versus grayscale computation, and then because the difference is small, leave further investigation of accuracy and efficiency to cited literature. For scale, one aspect of this attribute is covered by its interdependency in the resolution experiments. However another aspect, the relationship of scale to model levels (equation \ref{eqn:scaleFltrs}), we leave to future work. This is because scale is less easily separated from particular applications -- specifically the size range of objects
in an application -- than for other attributes. Our plan is to investigate this in a more application-oriented paper in the future.

\subsection{Similarity Metric Efficiency}
\label{sec:similarityEfficiency}
We could discover the effects of data attributes by training and inference testing all combinations of attribute values. For $N_{CL}$ classes, binary classification of pairs would require $N_{CL} \choose 2$ training and inference operations. In comparison, for similarity metrics, we need to train once for all combinations of binary classifications. During testing, the similarity model (SM) caches the latent space, thus only indices of each instance need to be paired with other instances to obtain their cosine similarities.

For the CIFAR10 dataset for example, there is a one-time task of feeding all test images into SM, which takes 0.72 seconds, and then caching, which takes 0.63 seconds. In contrast, we feed each image into the CNN to obtain its prediction in order to calculate its accuracy in the conventional pipeline.
 We show the runtime in Table ~\ref{tab:sim_eff}~.


 \begin{table}[h]
  \centering
\begin{tabular}{@{}l p{50pt} p{40pt}}
    \toprule
    Model & $t_{train}$ (s/epoch) & $t_{test}$ (s/pair) \\
    \midrule
    VGG19 & 0.69 & \hspace{2pt} 4.49 \\
    EfficientNet-B0 & 3.13 & 21.82 \\
    MobileNetV2 & 2.19 & 15.05 \\
    \textbf{Similarity Metrics} & 3.31 & \hspace{2pt} \textbf{0.76} \\
    \bottomrule
  \end{tabular}
  \caption{Results of runtime comparison of Similarity Metrics and existing models. s: second, pair: all pair of instances in two classes.}
  \label{tab:sim_eff}
\end{table}

\vspace{-6pt}
\subsection{Number of Classes, $N_{Cl}$}
\label{sec:Exp_numClasses}

It is well known that accuracy is reduced when more classes are involved since more visual features are needed for a CNN to learn in the feature extractor backbone, as well as more complex decision boundary to learn in the fully connected layers for classification. However, we perform experiments here, first to confirm this relationship with test data, but secondly to gain any extra insight into the relationship between number of classes and accuracy.

\begin{figure*}[h]
    \centering
        \includegraphics[width=0.33\linewidth]{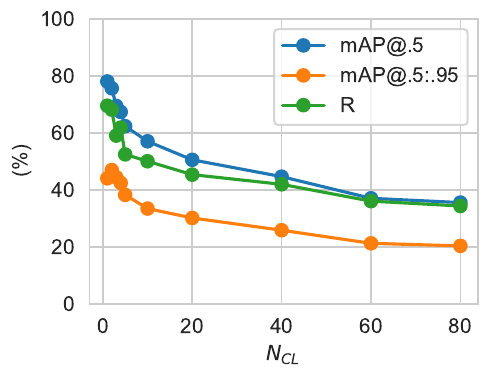}
        \includegraphics[width=0.33\linewidth]{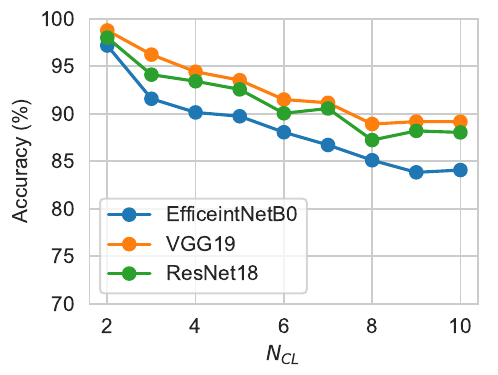}
        \includegraphics[width=0.33\linewidth]{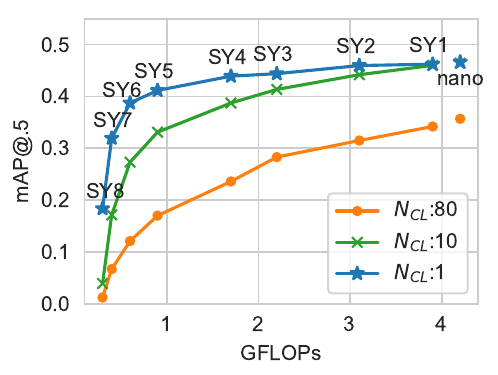}
    \caption{Overall relationship of performance and $N_{CL}$. (Left) Object detection accuracy and recall (R) decrease for the YOLOv5-nano when the number of classes is increased from 1 to 80. (Middle) CIFAR-10 Image classification accuracy decreases for the classifiers tested when the number of classes per group is increased from 2 to 10. (Right) Accuracy plot for increasingly smaller models from YOLOv5-nano through 8 sub-YOLO models (SY1-8) and  class groupings of 1 ($N_{CL}$:1), 10 ($N_{CL}$:10), and 80 ($N_{CL}$:80).}
    \label{fig:n_cls_plots}
\end{figure*}

We performed three sets of experiments. The first was for object detection using the YOLOv5-nano ~\cite{glenn_jocher_2021_5563715} backbone upon increasing-size class groupings of the COCO dataset ~\cite{lin2014microsoft}. Ten groups with $Cl$ of $\{1, 2, 3, 4, 5, 10, 20, 40, 60, 80\}$ were prepared. For each group, we trained a separate YOLOv5-nano model from scratch. As seen in Figure \ref{fig:n_cls_plots} (left), accuracy decreases with number of classes. An added insight is that the accuracy decrease is steep for very few classes, say 5-10 or fewer, and flattens beyond 10.

The second set of experiments was for image classification on the CIFAR-10 dataset. With many fewer classes in CIFAR-10 ~\cite{krizhevsky2009learning} than COCO (10 versus 80), we expect to see how the number of classes and accuracy relate for this smaller range. We extracted subsets of classes --- which we call groups -- from CIFAR-10 with $N_{Cl}$ ranging from 2 to 9. For example, a group with $N_{Cl}=4$ might contain airplane, cat, automobile, and ship classes. We trained classifiers from scratch for each group.

Results of the image classification experiments are shown in Figure \ref{fig:n_cls_plots} (middle). The three classifiers used for testing, EfficientNet-B0, VGG19 \cite{vggNet2014}, and MobileNetV2 \cite{sandler2018mobilenetv2} showed the expected trend of accuracy reduction as $N_{Cl}$ per group increased. However, the trend was not as monotonic as might be expected. We hypothesized that this might be due to the composition of each group, specifically if the group classes were largely similar or dissimilar. This insight led to the experiments on class similarity in the next section.

The third set of experiments involved reducing model size for different numbers of classes. We prepared 90 class groupings extracted from the COCO minitrain dataset ~\cite{HoughNet}. There are 80 datasets for $N_{Cl}=1$, each containing a single class from 80 classes. There are 8 datasets for $N_{Cl}=10$ combining the 8 single-class datasets. The final dataset is the original COCO minitrain with $N_{Cl}=80$.

We scale YOLOv5 layers and channels with the depth and width multiples already used for scaling the family between nano and x-large. Starting with depth and width multiples of 0.33 and 0.25 for YOLOv5-nano, we reduce these in step sizes of 0.04 for depth and 0.03 for width. In this way, we design a monotonically decreasing sequence of sub-YOLO models denoted as SY1 to SY8. We train each model separately for each of the six datasets.

Results of sub-YOLO detection are shown in Figure \ref{fig:n_cls_plots} (right). There are three lines where each point of $mAP@.5$ is averaged across all models in all datasets for a specific $N_{Cl}$. An overall trend is observed that fewer-class models (upper-left blue star) achieve higher efficiency than many-class models. Another finding of interest here is that, whereas the accuracies for 80 classes drops steadily from the YOLOv5-nano size, accuracy for 10 classes is fairly flat down to SY2, which corresponds to a 36\% computation reduction, and for 1 class down to SY4, which corresponds to a 72\% computation reduction.


\subsection{Color, $N_{Co}$}
\label{sec:Exp_color}
Because reduction from color to grayscale only affects the number of multiplies in the first layer, the efficiency gain depends upon the model size. For a large model such as VGG-19, percentage efficiency gain will be much smaller than for a small model such as EfficientNet-B0, as shown in Table \ref{tab:colorComparison}. However, even for small networks, the effect of reducing from color to grayscale processing is small relative to effects of other attributes, so we perform experiments on these more impactful attributes. For further investigation of color computation, refer to previous work on this topic \cite{grayscaleBetter2016}.

\begin{table}[h]
  \centering
  \begin{tabular}{p{35pt}|p{28pt}|p{25pt}|p{27pt}|p{27pt}|p{20pt}}
  
  \toprule
  Model & \multicolumn{2}{|c|}{Layer-1} & \multicolumn{2}{|c|}{All Layers} & Ratio \\
  \cmidrule{2-6}
    & color & gray & color & gray & [\%]\\
   \midrule
    VGG-19 & 1835.01 & 655.36 & 399474 & 398295 & 99.7 \\
    EN-B0 & 884.7 & 294.9 & 31431 & 30841 & 98.1 \\
   \midrule
  \end{tabular}
  \caption{ Difference between color and grayscale computation [kFLOPS] for the large VGG-19 classifier and much smaller EfficientNet-B0 classifier. Ratio is grayscale-to-color computation for all layers. 
  }
  \label{tab:colorComparison}
\end{table}

\subsection{Intra- and Inter-Class Similarity, $S_1$, $S_2$}
\label{sec:Exp_intrasim}

\label{sec:Exp_intersim}

In equation \ref{eqn:EoC}, we hypothesized that accuracy is lower if inter-class similarity is higher. Table \ref{tab:group_sim} shows accuracy and inter-class similarity results for groups of 2 and 4 classes from the CIFAR-10 dataset that we have subjectively designated similar (S) or dissimilar (D). ``S4'' indicates a similar group of 4 classes, and ``D2'' indicates a dissimilar group of 2 classes. The results indicate that our subjective labels correspond to our hypothesis, and that our objective measure of inter-class similarity in equation \ref{eqn:S2} both corresponds to the subjective, and is consistent with the hypothesis.

\begin{table}[h]
  \centering
  \begin{tabular}{@{}p{12pt} p{71pt} | p{19pt} p{19pt} p{19pt} | p{19pt}}
    \toprule
    G & Class Label & EB0 & V19 & MV2 & I-CS \\
    \specialrule{.10em}{.05em}{.05em} 
    $S4$ & cat, deer, \newline  dog, horse & 0.84 & 0.86 & 0.76 & \underline{0.57} \\
    \midrule
    $D4$ & airplane, cat, \newline  automobile, ship & 0.91 & 0.94 & \underline{0.93} & 0.12 \\
    \specialrule{.10em}{.05em}{.05em} 
    $S2_{1}$ & deer, horse & \underline{0.92} & 0.94 & 0.89 & \textbf{0.61} \\
    $S2_{2}$ & automobile, truck & 0.91 & \underline{0.95} & \underline{0.93} & 0.56 \\
    \midrule
    $D2_{1}$ & airplane, frog & \textbf{0.98} & \textbf{0.98} & \textbf{0.96} & 0.11 \\
    $D2_{2}$ & deer, ship & \textbf{0.98} & \textbf{0.98} & \textbf{0.96} & 0.08 \\
    \bottomrule
  \end{tabular}
  \caption{Accuracies of three image classifiers (EB0, V19, and MV2) for class groupings (G), and Inter-Class Similarity (I-CS). S denotes a similar group while D denotes a distinguishable group. Highest accuracy is highlighted in bold while the second highest is denoted by an underscore.}
  \label{tab:group_sim}
\end{table}

\begin{figure*}[h]
  \centering
    \includegraphics[width=0.335\linewidth]{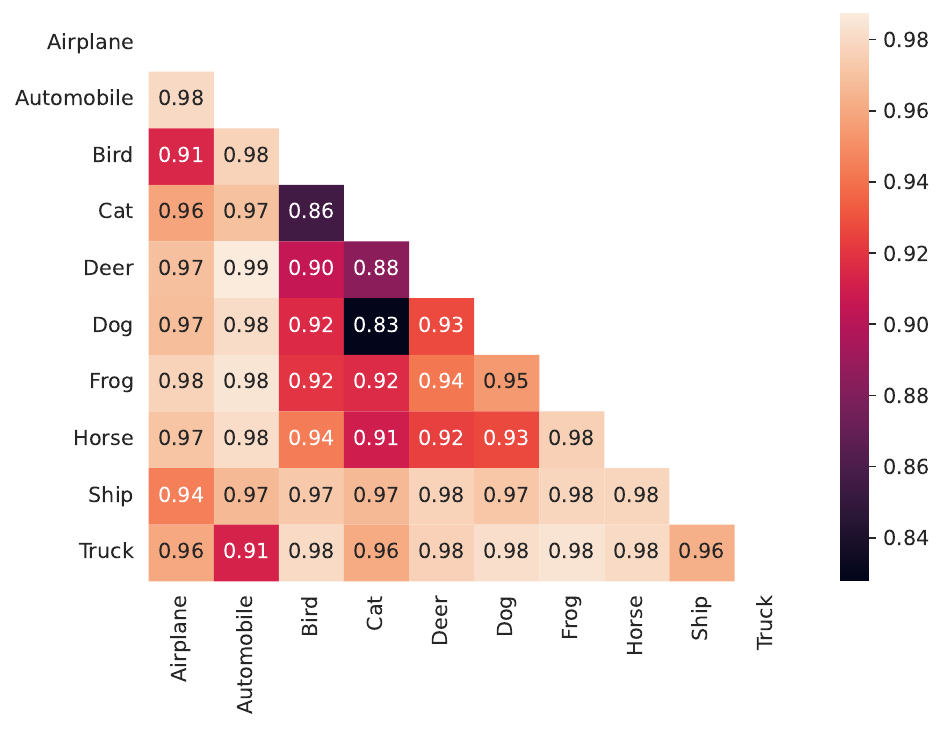}
    \includegraphics[width=0.335\linewidth]{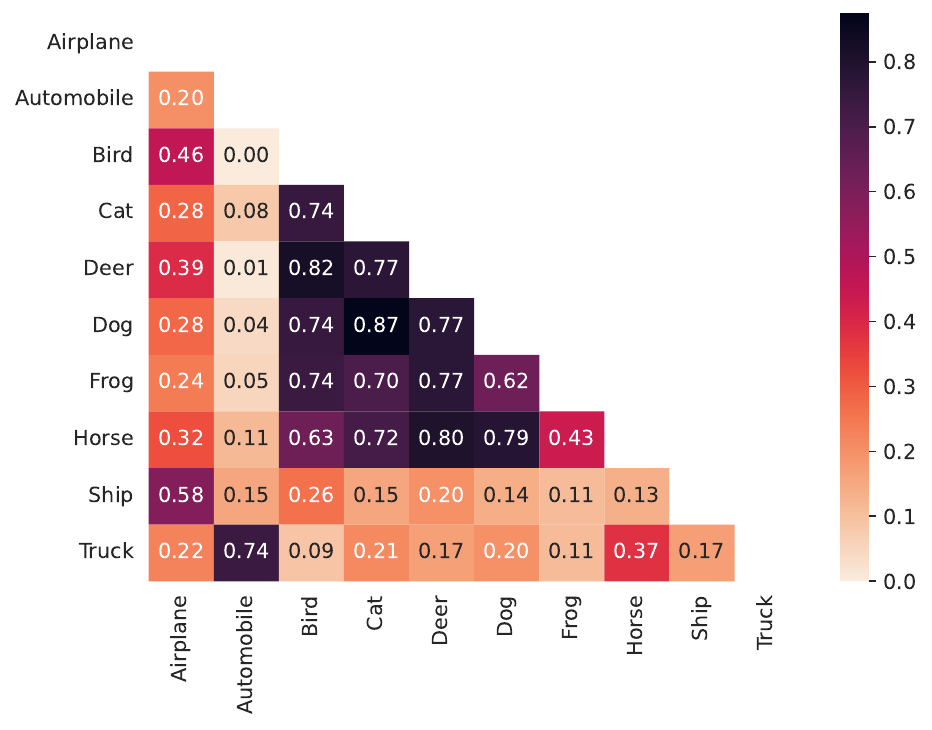}
    \includegraphics[width=0.28\linewidth]{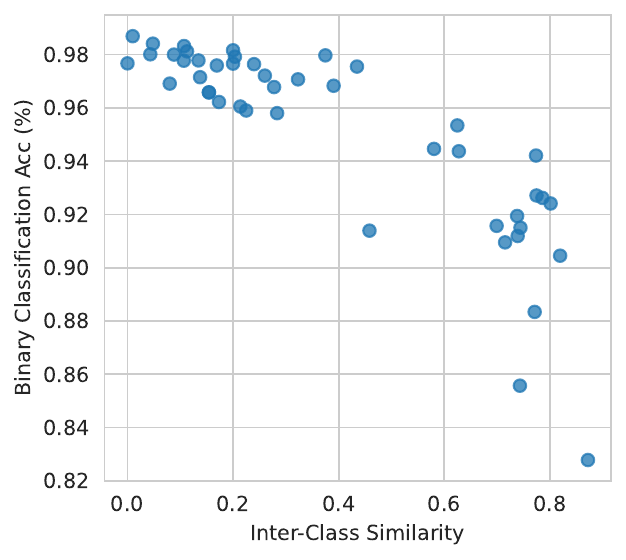}
  \caption{Matrices showing relationships among pairs of classes: (Left) binary classification accuracy matrix using EfficientNet-B0, (Middle) binary-class similarity matrix with $S_2$ metric, (Right) relationship between binary classification accuracy (left) and similarity scores (middle), plotted to show the high negative correlation between the two.}
  \label{fig:sim_matrix}
\end{figure*}

To further support the objectivity of our similarity measures, we compare similarity scores to accuracies for all pairwise classifications in CIFAR-10. The Pearson correlation coefficient between these two matrices is -0.77, showing strong correlation; it is negative because high accuracy corresponds with low similarity. All (similarity, accuracy) pairs are plotted on Figure \ref{fig:sim_matrix} (right) clearly indicating the inverse relationship between similarity and accuracy. Notably, the lowest similarity data point is for the (automobile, deer) pair, and the highest similarity data point is for the (cat, dog) pair.

\begin{table}[h]
  \centering
  \begin{tabular}{@{}p{13pt}|p{9pt}p{9pt}p{9pt}p{9pt}p{9pt}p{9pt}p{9pt}p{9pt}p{9pt}p{9pt}}
    \toprule
    $ID_{c}$ & 0 & 1 & 2 & 3 & 4 & 5 & 6 & 7 & 8 & 9 \\
    \midrule
    $S_{1}$ & 0.73 & 0.80 & 0.82 & 0.85 & 0.92 & 0.92 & 0.99 & 1.00 & 0.71 & 0.83 \\
    \bottomrule
  \end{tabular}
  \caption{Intra-class similarity for CIFAR10. The class numbers $ID_{c}$ correspond to the class names in Figure \ref{fig:sim_matrix}}
  \label{tab:intra_sim}
\end{table}

Intra-class similarity scores are shown for the same CIFAR-10 classes as in Table \ref{tab:intra_sim}. All classes show high similarity scores with some differences. One can subjectively justify these numbers by visual examination. Classes 6 and 7 (dog and frog) have the highest similarity scores because foreground and background are similar for most images. Classes 0 and 8 (airplane and ship) have lower similarity scores because there is more diversity in foreground object and background.



\subsection{Resolution and Scale, $R_E$ and $S_C$}
\label{sec:Exp_res}
\label{sec:Exp_scale}

\begin{figure}[h]
\centering
    \includegraphics[width=0.3\linewidth]{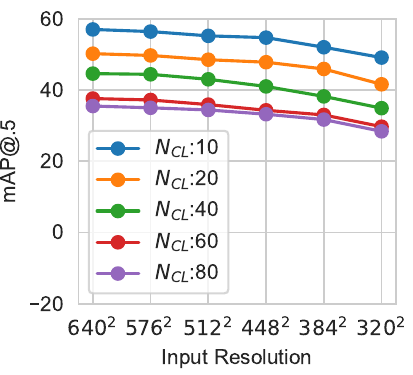}
    \hspace{40pt}
    \includegraphics[width=0.3\linewidth]{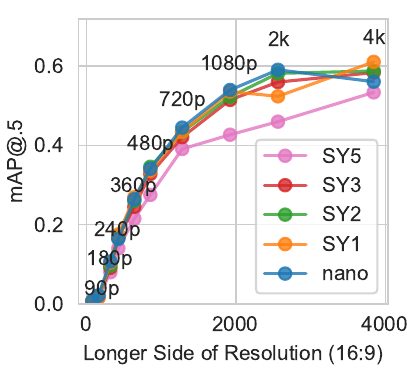}
   \caption{(Left) Effect of resolution on accuracy for YOLOv5-nano object detection on the COCO dataset, and (Right) effect on accuracy of YOLOv5 and sub-YOLO models for 80 classes of the PANDA 4k dataset ~\cite{wang2020panda}~.
   }
   \label{fig:YOLOv5_nano_coco_PANDA}
\end{figure}

In Table \ref{tab:applications}, the applications often downsize images to much smaller than the originals. Designers of these applications likely determined that accuracy didn't suffer when images were reduced to the degree that they chose. This is also the case for our experiment of YOLOv5 nano on the COCO dataset in Figure \ref{fig:YOLOv5_nano_coco_PANDA} (left). One can see that accuracies drop slowly for image reduction from $640^2$ to $448^2$ -- and further depending upon the degree of accuracy reduction that can be tolerated. Because scale drops with resolution, this plot includes the effects on accuracy reduction for both. In Figure \ref{fig:YOLOv5_nano_coco_PANDA} (right) we see the object detection performance almost flattens when reducing $R_{E}$ from 4k down to 1080p, and SY1-3 models' accuracies are very close to nano but with about half the required computation (4.2 GFLOPs for nano versus 2.2 GFLOPs for SY3.

We also verify by experiment that $R_{E}$ has a direct impact on inference runtime per image ~\cite{meng2020ar, yang2020resolution}. Experiments on YOLOv5 nano in the COCO validation set are conducted, and results are shown in Table \ref{tab:res_time}. One observation is that runtime almost doubles from 7.4ms for $1280^{2}$ to 13.4ms for $2560^{2}$ on a GPU, while it rises dramatically from 64.1 for $640^{2}$ to 4396.7 for $1280^{2}$ on a CPU.

\begin{table}[h]
  \centering
  \begin{tabular}{p{16pt}|p{13pt}p{13pt}p{13pt}p{13pt}p{23pt}p{23pt}p{24pt}}
    \toprule
    $R_{E}$ & $80^2$ & $160^2$ & $320^2$ & $640^2$ & $1280^2$ & $2560^2$ & $5120^2$ \\
    \midrule
    GPU & 6.4 & 6.2 & 6.3 & 6.4 & 7.4 & 13.4 & 48.7 \\
    CPU & 14.0 & 24.7 & 52.8 & 64.1 & 4396.7 & 4728.3 & 5456.9 \\
    \bottomrule
  \end{tabular}
  \caption{YOLOv5 runtimes [ms/img] for resolutions of the COCO validation set. Each number is averaged for 5 runs.}
  \label{tab:res_time}
\end{table}
\subsection{Robot Path Planning Application}
\label{sec:applRobot}
We briefly mention results of methods from this paper used for efficient model design for our own application. The application is to detect times, locations, and densities of human activity on a factory floor for robot path planning. Initial labelling comprised 5 object classes: 2 of humans engaged in different activities, and 3 of different types of robots. Similarity metrics guided a reduction to 2 classes enabling a smaller model with computation reduction of 66\% and accuracy gain of 3.5\%. See \cite{ogorman2022pathplan} for a more complete description of this application.

\section{Conclusion}
\label{sec:conclusion}

We conclude that for applications requiring lightweight CNNs, data attributes can be examined and adjusted to obtain more efficient models. We examined four independent data-side variables, and results from our experiments indicate the following ranking upon computation reduction. Resolution has the greatest effect on computation. Most practitioners already perform resolution reduction, but many simply to fit the model of choice. We show that, for small (few-class) applications the model size can be reduced (to sub-YOLO models) to achieve more efficiency with similar accuracy, and this can be done efficiently using similarity metrics. Number of classes is second in rank. This is dependent on the application, but our methods using similarity metrics enable the application designer to compare different class groupings efficiently. We showed that the choice of color or grayscale had a relatively small (1-2\%) effect on computation for small models. We don't rank scale because scale was only tested with interdependency to resolution.





\bibliographystyle{unsrt}  
\bibliography{main}

\end{document}